\documentclass{article}

\PassOptionsToPackage{numbers, compress}{natbib}
\usepackage{natbib}

\usepackage[preprint]{neurips_2024}

\usepackage[utf8]{inputenc} % allow utf-8 input
\usepackage[T1]{fontenc}    % use 8-bit T1 fonts
\usepackage{hyperref}       % hyperlinks
\usepackage{url}            % simple URL typesetting
\usepackage{booktabs}       % professional-quality tables
\usepackage{amsfonts}       % blackboard math symbols
\usepackage{nicefrac}       % compact symbols for 1/2, etc.
\usepackage{microtype}      % microtypography
\usepackage{xcolor}         % colors
\usepackage{amsmath}
\usepackage{amssymb}
\usepackage{algorithm}
\usepackage{algorithmic}
\usepackage{booktabs}
\usepackage{graphicx}
\usepackage{url}  % To handle URLs in the bibliography

\title{Well2Flow: Reconstruction of reservoir states from sparse wells using score-based generative models}

\author{%
  Shiqin Zeng\thanks{Equal contribution.} \\
  Georgia Institute of Technology \\
  \texttt{szeng44@gatech.edu}
  \And
  Haoyun Li\footnotemark[1] \\
  Georgia Institute of Technology \\
  \texttt{haoyun.li@gatech.edu}
  \And
  Abhinav Prakash Gahlot \\
  Georgia Institute of Technology \\
  \texttt{agahlot8@gatech.edu}
  \And
  Felix J. Herrmann \\
  Georgia Institute of Technology \\
  \texttt{felix.herrmann@gatech.edu}
}

\begin{document}

\maketitle

\begin{abstract}
This study investigates the use of score-based generative models for reservoir simulation, with a focus on reconstructing spatially varying permeability and saturation fields in saline aquifers, inferred from sparse observations at two well locations. By modeling the joint distribution of permeability and saturation derived from high-fidelity reservoir simulations, the proposed neural network is trained to learn the complex spatiotemporal dynamics governing multiphase fluid flow in porous media. During inference, the framework effectively reconstructs both permeability and saturation fields by conditioning on sparse vertical profiles extracted from well log data. This approach introduces a novel methodology for incorporating physical constraints and well log guidance into generative models, significantly enhancing the accuracy and physical plausibility of the reconstructed subsurface states. Furthermore, the framework demonstrates strong generalization capabilities across varying geological scenarios, highlighting its potential for practical deployment in data-scarce reservoir management tasks.
\end{abstract}

\section{Introduction}

Understanding and predicting subsurface flow is crucial for the sustainable management of natural resources and environmental systems. Recent advances in deep learning (DL) \cite{DeepLearningforSubsurfaceFlow} have opened new avenues in this field by enabling the development of fast and accurate surrogate models that capture the spatiotemporal dynamics of fluid states, such as gas saturation and reservoir pressure. However, the heterogeneity of geological formations limits the accuracy of subsurface characterization, reducing the adaptability of both surrogate models and traditional simulations. And integrating well log data into history matching remains challenging due to limited spatial coverage and model alignment complexity. To overcome these limitations, we propose a novel framework based on score-based generative models, in which the sampling process is conditioned on sparse well log measurements and constrained by physical laws expressed as partial differential equations (PDEs). By incorporating physical constraints and well log information directly into the generative process, our approach produces physically consistent reconstructions that align with both observational data and governing physics. Experimental results demonstrate that the proposed method can accurately reconstruct complex reservoir states, effectively leveraging limited well log data to capture fine-scale geological features and enhance predictive performance.

\section{Guided diffusion framework}
\subsection{Diffusion models}
Recent advancements in diffusion models have demonstrated their capacity to learn robust implicit priors from complex data distributions. For example, score-based generative modeling \cite{song2021scorebasedgenerativemodelingstochastic} illustrated that by matching the gradient of the log density (\( \nabla_x \log p(x) \)), these models can effectively capture intricate structures inherent in the data. Diffusion posterior sampling (DPS) \cite{chung2024diffusionposteriorsamplinggeneral} highlights the potential to incorporate the logarithmic likelihood gradient to guide the denoising process, allowing the approximation of posterior samples based on sparse observations. Additionally, DiffusionPDE    \cite{huang2024diffusionpdegenerativepdesolvingpartial} demonstrates how learned generative priors can be applied to solve a wide range of PDEs under partial observations, further emphasizing the potential of diffusion models to integrate with physical modeling in real-world applications. In our cases, we train the diffusion model on the joint distribution of full permeability $K$ and saturation $S$ of the final states based on the JutulDarcy.jl
 \cite{jutuldarcy_ecmor_2024,li_2025_14927938}. During inference, the model begins from Gaussian noise and iteratively denoises the samples, guided by both sparse well log observations and the governing PDE constraints. This physics-informed generative approach enables the accurate reconstruction of both permeability and saturation fields, ensuring consistency with the observed data and underlying physical laws.

\subsection{Observation guidance }
Elucidating the design space of diffusion-based generative models (EDM) \cite[]{karras2022elucidatingdesignspacediffusionbased} proposed to learn the denoiser function \( D(x; \sigma) \) that effectively approximates the score function such that \( \nabla_x \log p(x; \sigma(t))  =  \frac{D(x; \sigma(t)) - x}{\sigma^2(t)} \). This equation is then applied to the probability flow Ordinary Differential Equation (ODE), which provides a deterministic trajectory for sample generation \cite[]{song2021scorebasedgenerativemodelingstochastic}. At each timestep \( t \), the denoised sample \( x \) follows:
\begin{equation}
    \frac{dx}{dt} = -\sigma'(t)\sigma(t) \left( \frac{D(x; \sigma(t)) - x}{\sigma^2(t)} \right)
\label{eq1}
\end{equation}
Here \(-\sigma'(t)\sigma(t)\)  serves as a time-dependent scaling factor for the score function,  controlling the rate and stability of the sample generation process. The inpainting problem in DPS \cite[]{chung2024diffusionposteriorsamplinggeneral} is well-suited for reservoir reconstruction, where a complete permeability-saturation field is inferred from limited observed well log data. To incorporate the guidance of well log information \( y \), DPS modifies the probability flow ODE by adding a likelihood guidance term: 
\begin{equation}
    dx = -\dot{\sigma}(t) \sigma(t) (\nabla_x \log p(x; \sigma(t)) + \nabla_x \log p(y | x; \sigma(t)))dt
\end{equation}
Applying Bayes' rule, the posterior distribution of the permeability and saturation field given well log observations (\(y\)) is 
\( p(x | y) \propto p(x) p(y | x) \), and the posterior at each denoising step \( i \) is approximated as:
\begin{equation}
    p(x_i | y) \propto p(x_i) p(y_i | x_i) \dots p(x_1) p(y_1 | x_1)
\end{equation}
Under the assumption that measurement errors follow a gaussian distribution, the log-likelihood gradient is approximated as:
\begin{equation}
    \nabla_x \log p(y | x_i; \sigma(t_i)) \approx -\frac{1}{\delta^2} \nabla_{x_i} \| y - Mx_i \|^2_2
\end{equation}
Guided by the observation data, the gradient direction of the diffusion is therefore:
\begin{equation}
    \nabla_{x_i} \log p(x_i | y) \approx s(x_i) - \zeta_{\text{obs}} \nabla_{x_i} \| y - M x_i \|_2^2
\label{obs}
\end{equation}
where \( s(x) = \nabla_x \log p(x) \) is the score function learned by the diffusion model, \( M \) is a 2D binary mask indicating the coordinates
 of well locations, and the second term enforces data consistency with the well log observations.

\subsection{Physical guidance}

\begin{figure*}[h]
    \centering
    \includegraphics[width=1.0\textwidth]{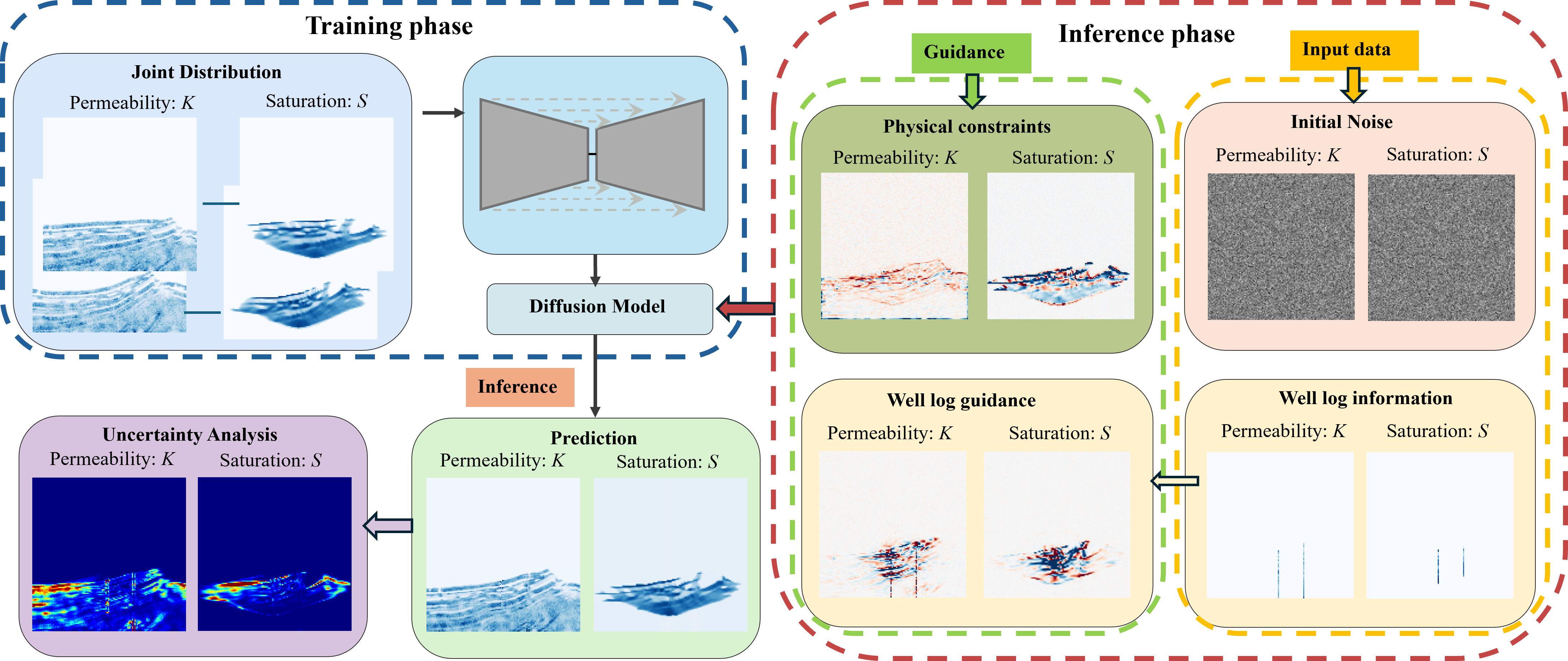}
    \caption{Work flow}
    \label{fig:flow}
\end{figure*}

We use Geological Carbon Sequestration (GCS) as a case study to evaluate the proposed methodology. The CO\textsubscript{2} injection process is constrained by phase mass conservation and Darcy’s law, formulated as:
\begin{equation}
    \nabla \cdot (\rho \vec{v}) = q,
\end{equation}
\begin{equation}
    \vec{v} = -\frac{K}{\mu} (\nabla P -  g \nabla z),
\end{equation}
where \( \rho \) is the CO\textsubscript{2} mass density, \( \vec{v} \) is the fluid flow velocity, and \( q \) is the source term representing CO\textsubscript{2} injection or production. Here, \( \mu \) is the CO\textsubscript{2} viscosity, \( P \) is the pressure, \( g \) is the gravitational acceleration, and \( z \) is the depth. To couple \( P \) and \( S \) based on the Brooks-Corey model, we can rewrite the equation as:
\begin{equation}
    h = -\nabla \cdot (K \nabla S) - q= 0
\end{equation}
Once the diffusion model \( D_{\theta} \) is trained, the physical constraint \( \nabla h = 0 \) can be enforced during the denoising step \( i \) \cite{huang2024diffusionpdegenerativepdesolvingpartial} by adding the PDE loss term:
\begin{equation}
    \nabla_{x_i} \log p(x_i | y, h) \approx s(x_i) - \zeta_{\text{obs}} \nabla_{x_i} \| y - M x_i \|_2^2  - \zeta_{\text{pde}} \nabla_{x_i} \| h(x_i)  \|_2^2.
\end{equation}
The details of the sampling process are outlined in Algorithm~\ref{alg:obs_pde_diffusion1}, while the framework of the guided diffusion is illustrated in Fig.~\ref{fig:flow}.

% First figure spanning both columns

\begin{algorithm}
\caption{Sampling process with well log observation and physical information guidance}
\label{alg:obs_pde_diffusion1}
\begin{algorithmic}[1]
\REQUIRE trained diffusion model $D_\theta(x; \sigma)$, denoising steps $N$, $\sigma(t_i)_{i \in \{0\ldots, N-1\}}$, well locations  $M$, well log  observation  $y$, physical constraints $h$, weights $\zeta_{\text{obs}}$, $\zeta_{\text{pde}}$
\STATE Sample $x_0 \sim \mathcal{N}(0,  \sigma(t_0)^{2} I)$ \hfill $\triangleright$ Generate the initial sampling noise
\FOR{$i \in \{0, \ldots, N-1\}$}
    \STATE $\hat{x}_i \leftarrow D_\theta(x_i; \sigma(t_i))$ 
    \STATE $d_i \leftarrow (x_i - \hat{x}_i) / \sigma(t_i)$
    \STATE $x_{i+1} \leftarrow x_i + (\sigma(t_{i+1}) - \sigma(t_i)) d_i$ \hfill $\triangleright$ Euler step from $\sigma(t_i)$ to $\sigma(t_{i+1})$ in Eq.~\eqref{eq1}
    \\ $\triangleright$ Second correction using trapezoidal rule
    \IF{$\sigma(t_{i+1}) \neq 0$} 
        \STATE $\hat{x}_{i} \leftarrow D_\theta(x_{i+1}; \sigma(t_{i+1}))$
        \STATE $d_i' \leftarrow (x_{i+1} - \hat{x}_{i}) / \sigma(t_{i+1})$
        \STATE $x_{i+1} \leftarrow x_i + (\sigma(t_{i+1}) - \sigma(t_i)) \left( \frac{1}{2} d_i + \frac{1}{2} d_i' \right)$
    \ENDIF

\STATE $L_{\text{obs}} \leftarrow  \left\| y - M\hat{x}_i \right\|_2^2$ \hfill $\triangleright$ Well log information guidance
\STATE $L_{\text{pde}} \leftarrow \left\|h(\hat{x}_i) \right\|_2^2$ \hfill $\triangleright$ Physical constraints
\STATE $x_{i+1} \leftarrow x_{i+1} - \zeta_{\text{obs}} \nabla_{x_i} L_{\text{obs}} - \zeta_{\text{pde}} \nabla_{x_i} L_{\text{pde}}$ \hfill $\triangleright$ Guide sampling
\ENDFOR
\RETURN $x_N$ \hfill $\triangleright$ Return the reconstruction data
\end{algorithmic}
\end{algorithm}

\section{Experiments}
\subsection{Results comparison}
\begin{figure*}[h]
    \centering
    \includegraphics[width=0.9\textwidth]{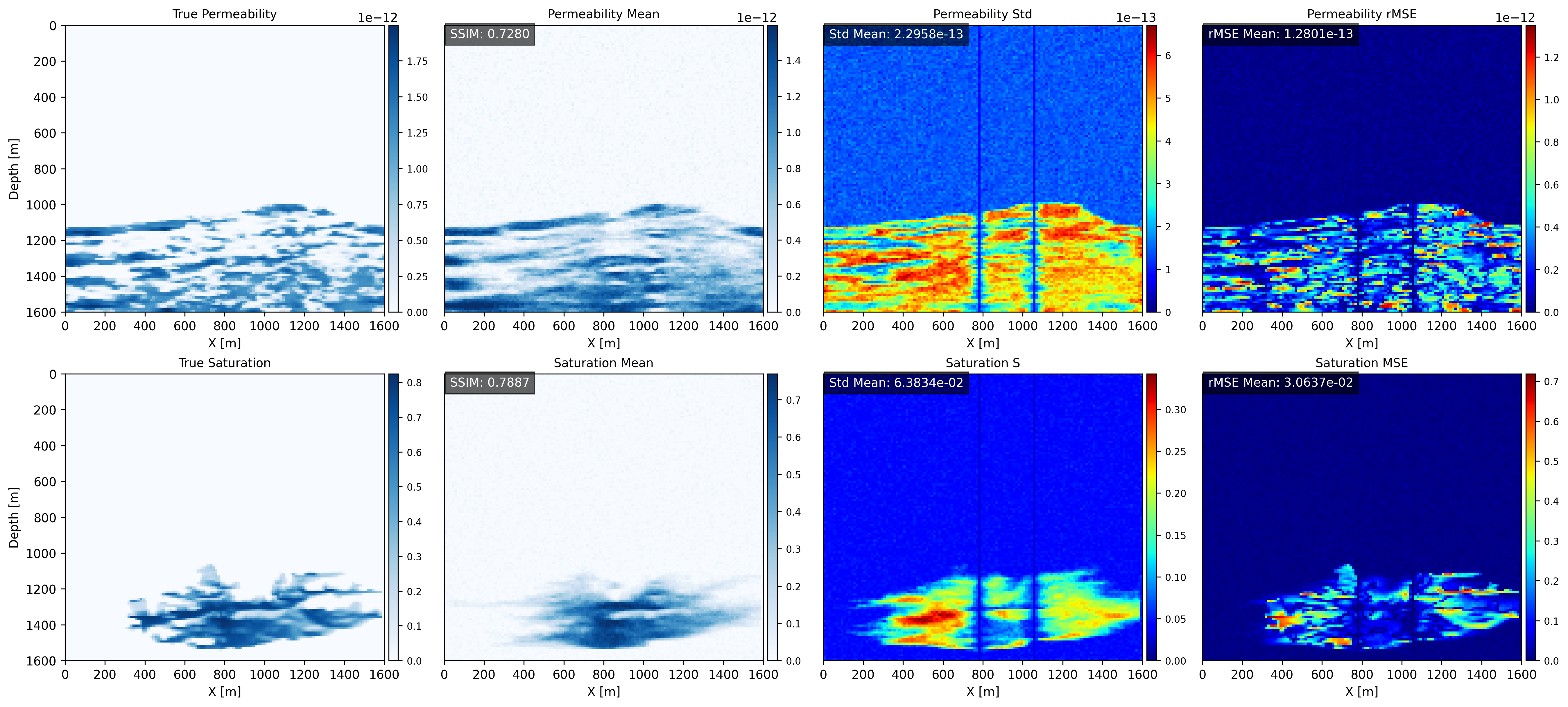}
    \caption{Uncertainty analysis based on two well points.}
    \label{fig:uncer2wells}
\end{figure*}

\begin{figure}[h]
    \centering
    \includegraphics[width=0.8\textwidth]{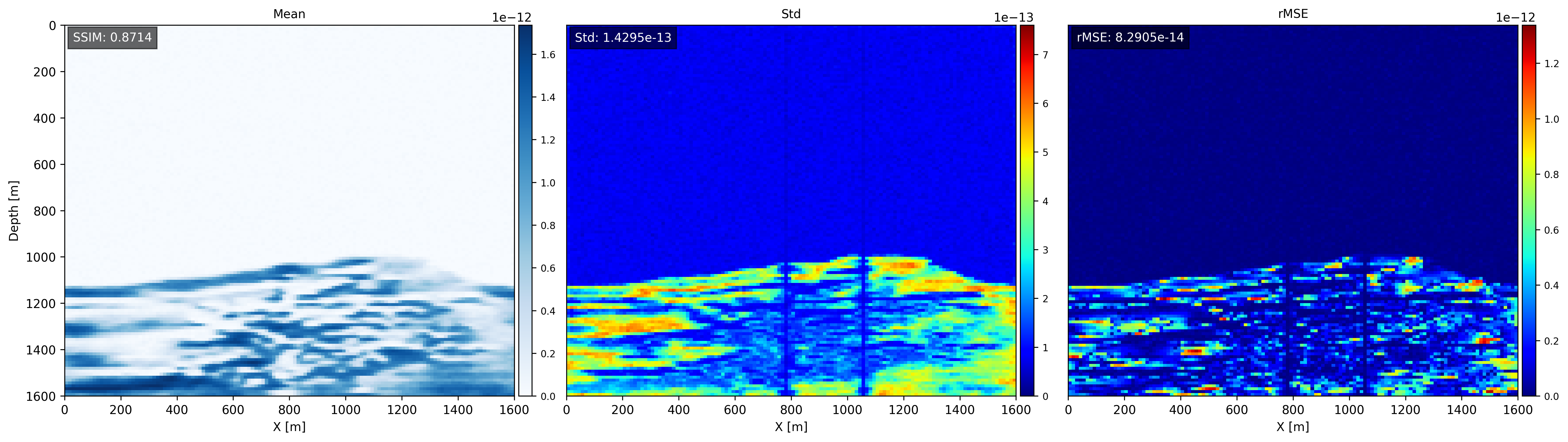}
    \caption{Permeability inversion with saturation and well guidance}
    \label{fig:forward}
\end{figure}

\begin{figure}[h]
    \centering
    \includegraphics[width=0.8\textwidth]{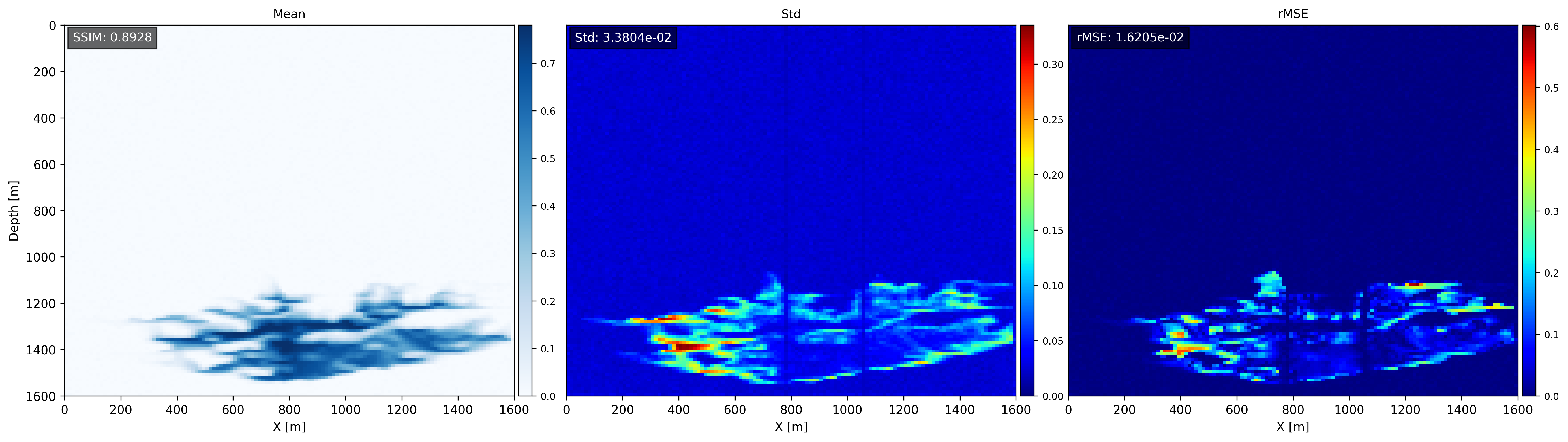}
    \caption{Saturation prediction with permeability and well guidance}
    \label{fig:inverse}
\end{figure}

With two well log data points located at horizontal distances of 775\,\(m\) and 1050\,\(m\), corresponding to the injection and production wells respectively, we conduct a series of experiments to evaluate the impact of different input configurations on reconstructing unseen data. The quantitative results are summarized in Table~\ref{tab:results}, while the qualitative results are presented in Fig.~\ref{fig:uncer2wells} through Fig.~\ref{fig:inverse}. Compared to the baseline scenario using only two wells, the reconstruction significantly improves in terms of both Root Mean Squared Error (rMSE) and Structural Similarity Index Measure (SSIM) when complete permeability or saturation information is provided, which indicate the score-based generative model learned the implicity mapping correctly,  The guidance from well log data further enhances reconstruction performance. These improvements are in addition to the learned mapping within the diffusion model, where well log information directly contributes to optimizing the permeability and saturation loss functions, offering more localized and precise supervision for permeability reconstruction. As shown in Table~\ref{tab:results}, the inclusion of well log data consistently boosts both SSIM and rMSE. Furthermore, incorporating physical constraints through PDE-based regularization yields reconstructed fields that are more physically consistent.

\begin{table}[h]
    \centering
    \caption{Results comparison}
    \resizebox{0.8\textwidth}{!}{  % Adjust table width
    \begin{tabular}{|c| c c| c c|} 
        \hline
        Input data & K rMSE & K SSIM & S rMSE & S SSIM \\ 
        \hline
        \hline
        Two wells  & 1.280e-13 &  0.7280 & 0.0306 & 0.7887 \\ 
        \hline
        Full K  & - & - & 0.0238 ↓ & 0.8537 ↑ \\ 
        \hline
        Full K + Two wells & - & - & \textbf{0.0164} ↓ & \textbf{0.8981} ↑ \\ 
        \hline
        Full S & 9.4841e-14 ↓ & 0.8534  ↑ & - & - \\ 
        \hline
        Full S + Two wells & \textbf{8.9682e-14} ↓ & \textbf{0.8637} ↑ & - & - \\ 
        \hline
    \end{tabular}
    }  % End resizebox
    \label{tab:results}
\end{table}


\begin{figure*}[h]
    \centering
    \includegraphics[width=0.9\textwidth]{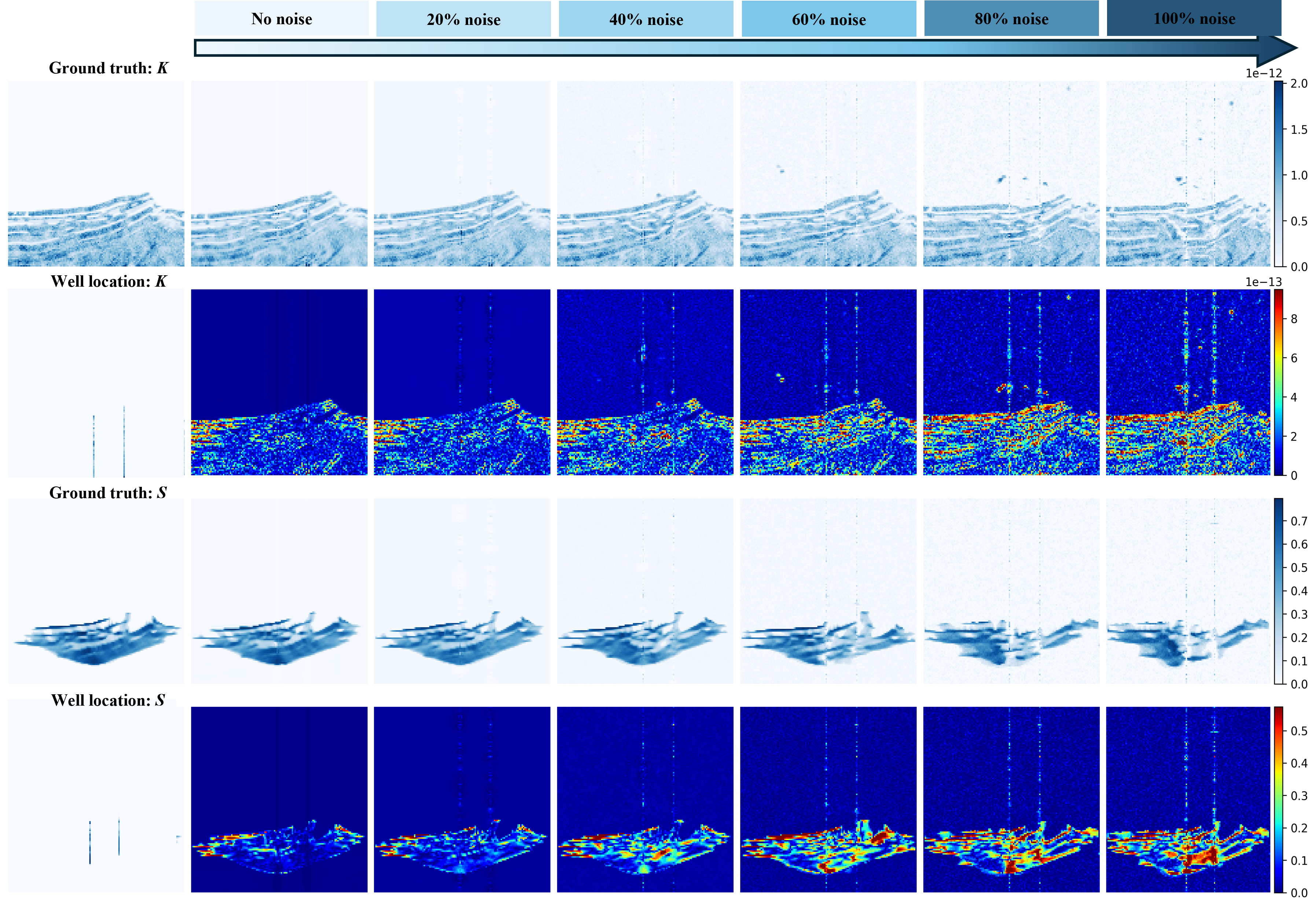}
    \caption{Ablation study: The first and third rows show the ground truth and generated permeability (\(K\)) and saturation (\(S\)) with different noise levels at well locations. The second and fourth rows display the corresponding root mean squared errors (rMSE).}
    \label{fig:ablation}
\end{figure*}

To further investigate the reconstruction ability of the model, an uncertainty analysis is conducted by aggregating results from 50 generated samples to compute the mean, standard deviation (std), and rMSE for both permeability and saturation fields,  as shown in Fig.~\ref{fig:uncer2wells}. The std maps reveal regions of high uncertainty, which closely correspond to areas with higher rMSE values. This relationship indicates that the uncertainty derived from the sample distribution effectively reflects potential model errors. Specifically, regions of higher uncertainty are located farther from the well, while lower uncertainty is associated with areas in closer proximity, guided by well information.

\subsection{Observation noise sensitivity study}
To evaluate the reliability of well log information in guiding the generation process, an ablation study is conducted, as shown in Fig.~\ref{fig:ablation}. Gaussian noise is added according to the mean \((\mu_k, \mu_s)\) and variance  \((\sigma_k^2, \sigma_s^2)\) of the ground truth permeability (\(k_{GT}\)) and saturation (\(s_{GT}\)), formulated as:  
\begin{align}
\begin{pmatrix} k_{well} \\ s_{well} \end{pmatrix} &\leftarrow M\left( \alpha \cdot \begin{pmatrix} \mathcal{N}(\mu_k, \sigma_k^2) \\ \mathcal{N}(\mu_s, \sigma_s^2) \end{pmatrix}  + (1-\alpha) \cdot \begin{pmatrix} k_{GT} \\ s_{GT} \end{pmatrix} \right) 
\end{align}
where \(\alpha\) is a noise weighting factor ranging from 0 to 1. As shown in Fig.~\ref{fig:ablation}, when the noise level increases to 40\%, the denoised well log information remains effective in guiding the reconstruction of permeability and saturation. This suggests that the model can still extract useful guidance from the well log data for reconstruction even in the presence of noise. However, when the noise level reaches 80\%, the reconstruction of permeability and saturation is severely corrupted, indicating that well log information becomes crucial for accurate generation. This ablation study demonstrates that while the model is able to learn informative patterns to guide the reconstruction process, it is vulnerable to high levels of noise. When the well log data is heavily corrupted, the model's ability to accurately generate permeability and saturation is compromised, highlighting the importance of correct well log information and the need for additional guidance in noisy environments.

\section{Conclusion and discussion}

This research introduces a novel framework for integrating well log data into a score-based generative model to reconstruct permeability and saturation fields in reservoir engineering. Rather than learning a direct mapping \( S = f(K) \), we train the model on the joint distribution \( x = (K, S) \), enabling it to implicitly capture the underlying relationship between reservoir characteristics and fluid states. This joint formulation allows permeability and saturation to mutually guide each other’s generation, leveraging multi-source well log data while naturally incorporating physical constraints through partial differential equations (PDEs). The framework enables flexible inference: with full permeability, it acts as a forward surrogate for saturation prediction; with full saturation, it performs inversion to estimate permeability. In both cases, well log data improves accuracy and reduces reconstruction error. This flexibility highlights the value of well log data in enabling both forward predictions and inverse reconstructions, offering new opportunities for data-driven solutions in reservoir modeling.

By integrating multimodal information of well logs (permeability and saturation), the proposed framework enhances both the accuracy and physical consistency of the reconstructed reservoir properties. The complementary nature of these data sources enables mutual guidance during generation, allowing the model to better capture the complex relationships between geological structures and fluid behavior. The noise sensitivity studies indicate that incorporating additional conditioning modalities is necessary to enhance model performance. Seismic imaging is a promising option, as reverse-time migration (RTM) can capture subsurface velocity variations that correlate with permeability. Furthermore, differences between baseline and monitoring seismic data can reveal the evolution of the CO\textsubscript{2} plume ~\cite{zeng2024ML4SEISMICepu}. Building on this potential, future work will focus on integrating seismic images alongside well log data and pressure measurements to further improve the reconstruction of permeability and saturation fields.

\section{Acknowledgments}
 This research was carried out with the support of Georgia Research Alliance and partners of the ML4Seismic Center.

\bibliography{main}
\end{document}